\documentclass[11pt,a4paper]{article}
\usepackage[hyperref]{acl2021}
\usepackage{times}
\usepackage{latexsym}
\usepackage{bbm}
\usepackage{booktabs}       
\usepackage{amsfonts}
\usepackage{nicefrac}
\usepackage{microtype}   
\usepackage{lipsum}
\usepackage{graphicx}
\usepackage{verbatim}
\usepackage{multirow}
\usepackage{multicol}
\usepackage{color}
\usepackage{latexsym}
\usepackage{graphicx}
\usepackage{xcolor}
\usepackage{colortbl,booktabs}
\usepackage{tabu}
\usepackage{float}
\usepackage{amsmath}
\usepackage{wrapfig}
\usepackage{multirow}
\usepackage{makecell}
\usepackage{caption}

\usepackage{microtype}

\aclfinalcopy 

\title{X-PuDu at SemEval-2022 Task 7: \\ A Replaced Token Detection Task Pre-trained Model with Pattern-aware Ensembling for Identifying Plausible Clarifications}

\author{
    Junyuan Shang$^1$,~~Shuohuan Wang$^1$,~~Yu Sun$^1$, \\ 
    \textbf{Yanjun Yu$^2$,~~Yue Zhou$^2$,~~Li Xiang$^2$} and \textbf{Guixiu Yang$^2$} \\
    \normalsize $^1$ Baidu Inc., China\\
    \normalsize $^2$ Shanghai Pudong Development Bank, China\\
    \small \{\texttt{shangjunyuan, wangshuohuan, sunyu02\}@baidu.com} \\
    \small \{\texttt{yuyj6, zhouy93, xiangl3, yanggx1\}@spdb.com.cn}
}

\date{}

\begin{document}
\maketitle
\begin{abstract}
This paper describes our winning system on SemEval 2022 Task 7: \textit{Identifying Plausible Clarifications of
Implicit and Underspecified Phrases in Instructional Texts}. A replaced token detection pre-trained model is utilized with minorly different task-specific heads for SubTask-A: \textit{Multi-class Classification} and SubTask-B: \textit{Ranking}. Incorporating a pattern-aware ensemble method, our system achieves a 68.90\% accuracy score and 0.8070 spearman's rank correlation score surpassing the 2nd place with a large margin by 2.7 and 2.2 percent points for SubTask-A and SubTask-B, respectively. Our approach is simple and easy to implement, and we conducted ablation studies and qualitative and quantitative analyses for the working strategies used in our system.
\end{abstract}

\section{Introduction}\label{sec:intro}
\begin{figure}[!t]
	\centering
	\includegraphics[width=0.45\textwidth]{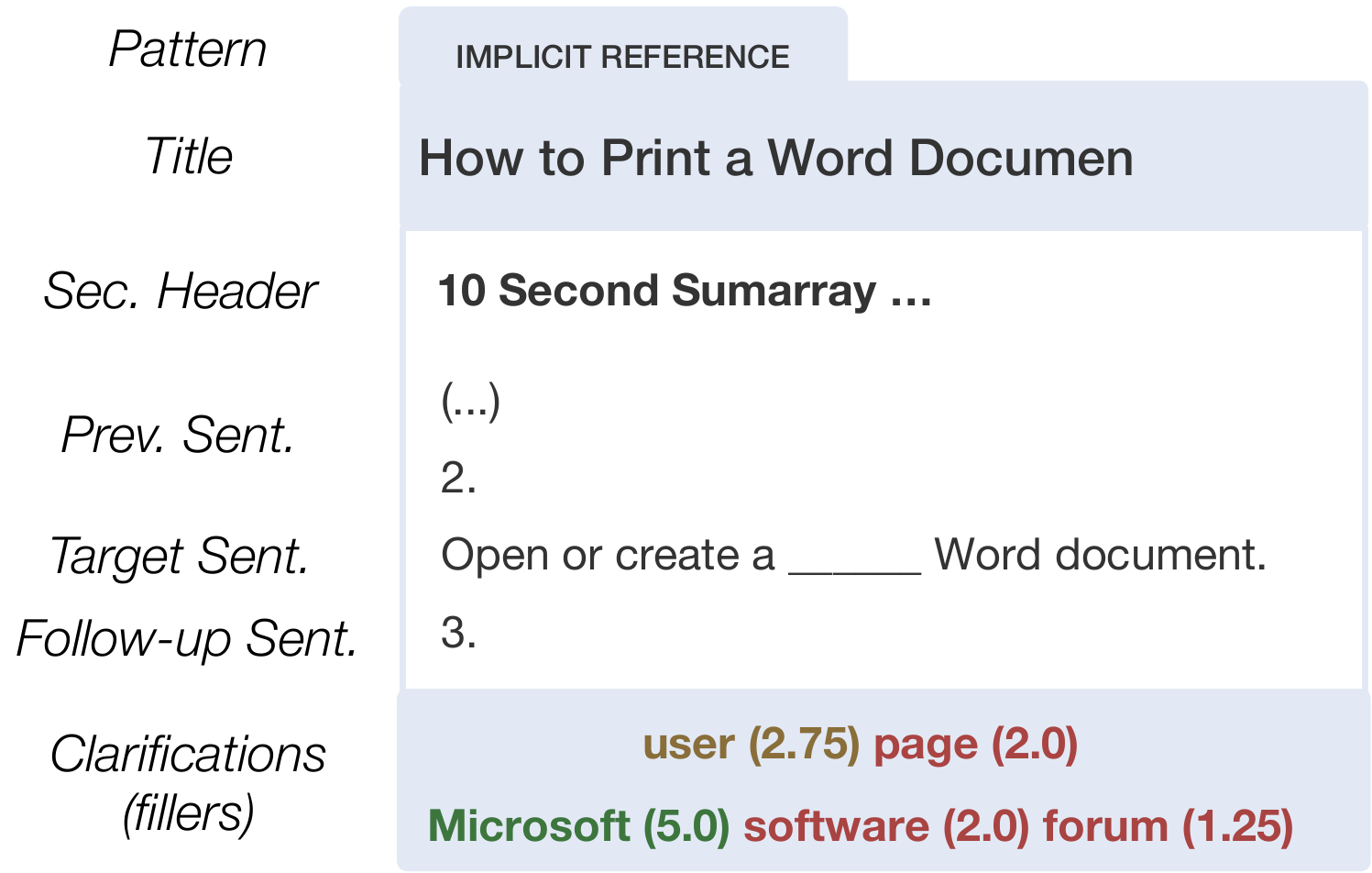}
	\caption{An example randomly chosen from the dev set. Each sample is associated with five clarifications labeled ( \textcolor[RGB]{60,118,60}{\texttt{PLAUSIBLE}}, \textcolor[RGB]{141,114,66}{\texttt{NEUTRAL}} or \textcolor[RGB]{178,99,101}{\texttt{IMPLAUSIBLE}}) and scored on a scale from 1.0 to 5.0.}\label{fig:ex}
\end{figure}

The Internet's ever-increasing size has made it easy to find instructional texts such as articles in wikiHow\footnote{\url{https://www.wikihow.com/Main-Page}}, on almost any topic or activity. Regular revisions of these how-to manuals are necessary to ensure that instructions communicate the procedures required to attain a certain goal precisely. This shared task is introduced by ~\citet{roth-etal-2022-identifying}, whose intention is to find ways to improve instructional texts, evaluate to what extent current NLP systems are able to handle implicit, ambiguous, and underspecified language, and go beyond the surface form of a text and take multiple plausible interpretations into account. Thus, the proposed NLP systems should be capable of distinguishing between plausible and implausible clarifications of an instruction shown in Figure.~\ref{fig:ex}. 

The shared task consists of two subtasks:
\begin{itemize}
\item \textbf{SubTask-A: Multi-Class Classification}. The goal is to predict a class label (IMPLAUSIBLE, NEUTRAL, PLAUSIBLE) given the clarification\footnote{A clarification is a word/phrase that was inserted to specify information in the instruction.} and its context.
\item \textbf{SubTask-B: Ranking}. The goal is to predict the plausibility score on a scale from 1 to 5 given the clarification and its context.
\end{itemize}

In this paper, we describe our winning system for both subtasks. We built our system based on a replaced token detection (RTD) task pre-training model. The idea is that the replaced token detection task is similar to this shared task which focuses on distinguishing semantically similar words/phrases. To close the gap the model trained between the pre-training phase and fine-tuning phase, we reused the pre-trained language modeling head during fine-tuning on Task 7. Then, two-layers MLPs are applied on the mean-pooled hidden states of a clarification (filler) given the context. For subtask A, we utilized the cross-entropy loss for the multi-class classification. For subtask B, a sigmoid function was used to impose restrictions on the output of the system on a scale from 1 to 5. Finally, we trained multiple models and aggregated the predictions with a pattern-aware ensemble strategy. Our system achieved the best overall performance in the shared task with a 68.9\% accuracy score (subtask A) and 0.807 Spearman's rank correlation score (subtask B). The outcomes are promising for improving the clarification of instructional texts.

\section{Background}
Pre-trained models~\cite{devlin2018bert,liu2019roberta,sun2019ernie,clark2020electra} have achieved state-of-the-art results in various Natural Language Processing
(NLP) tasks. Recent works~\cite{raffel2019exploring, brown2020language, sun2021ernie} have shown that more generalization ability and superior performance can be achieved by pre-training models with billion or trillion parameters. Thus, we pursued the competitive pre-trained models such as DeBERTa~\cite{he2020deberta} and large-scale pre-trained models ERNIE~\cite{sun2021ernie} whose effectiveness has been validated in the standard GLUE~\cite{wang2018glue} and SuperGLUE benchmark~\cite{wang2019superglue}.

However, in our initial experiment, we found that the aforementioned models, though have promising results on sentence-level or paragraph-level tasks, failed to distinguish the word/phrase-level semantically similar clarifications (fillers) in a given context. We believe the failure is due to the way these models were trained using a masked language modeling (MLM) task in the pre-training phase. MLM aims to map tokens with similar semantics to the embedding space that are close to each other instead of distinguishing them.

Based on the above finding, we believe what we need is a discriminator~\cite{clark2020electra, he2021debertav3} pre-trained via a replaced token detection (RTD) task which is more aligned with this shared task. In RTD, the discriminator needs to determine if a corresponding token is either an original token or a token replaced by the generator. Formally, the loss function for the discriminator is as follows:
\begin{equation}
    \mathcal{L}_{RTD}= -\sum_{i} \log p\left(\mathbbm{1}\left(\tilde{x}_{i}=x_{i}\right) \mid \tilde{\mathbf{X}}, i\right)
\end{equation}
where $\tilde{\mathbf{X}}$ is the input sequence constructed by replacing masked tokens with plausible tokens sampled from a generator, and the indicator function $\mathbbm{1}(\cdot)$ distinguishes whether the plausible tokens are generated or the original ones.

\section{Method}
In this section, we will describe the strategies we used in our system in detail. In Section.~\ref{sec:basic-model}, the system is presented on how we formalize the data as input, basic modules, and task-specific design for each subtask. Then, we describe the optimization object for each subtask (see Section.~\ref{sec:opt-obj}). Finally, we introduce a pattern-aware ensemble strategy to further boost the performance beyond a normal ensembled model in Section.~\ref{sec:ensemble}.

\subsection{System Description}\label{sec:basic-model}
\begin{figure*}[!htb]
	\centering
	\includegraphics[width=0.9\textwidth]{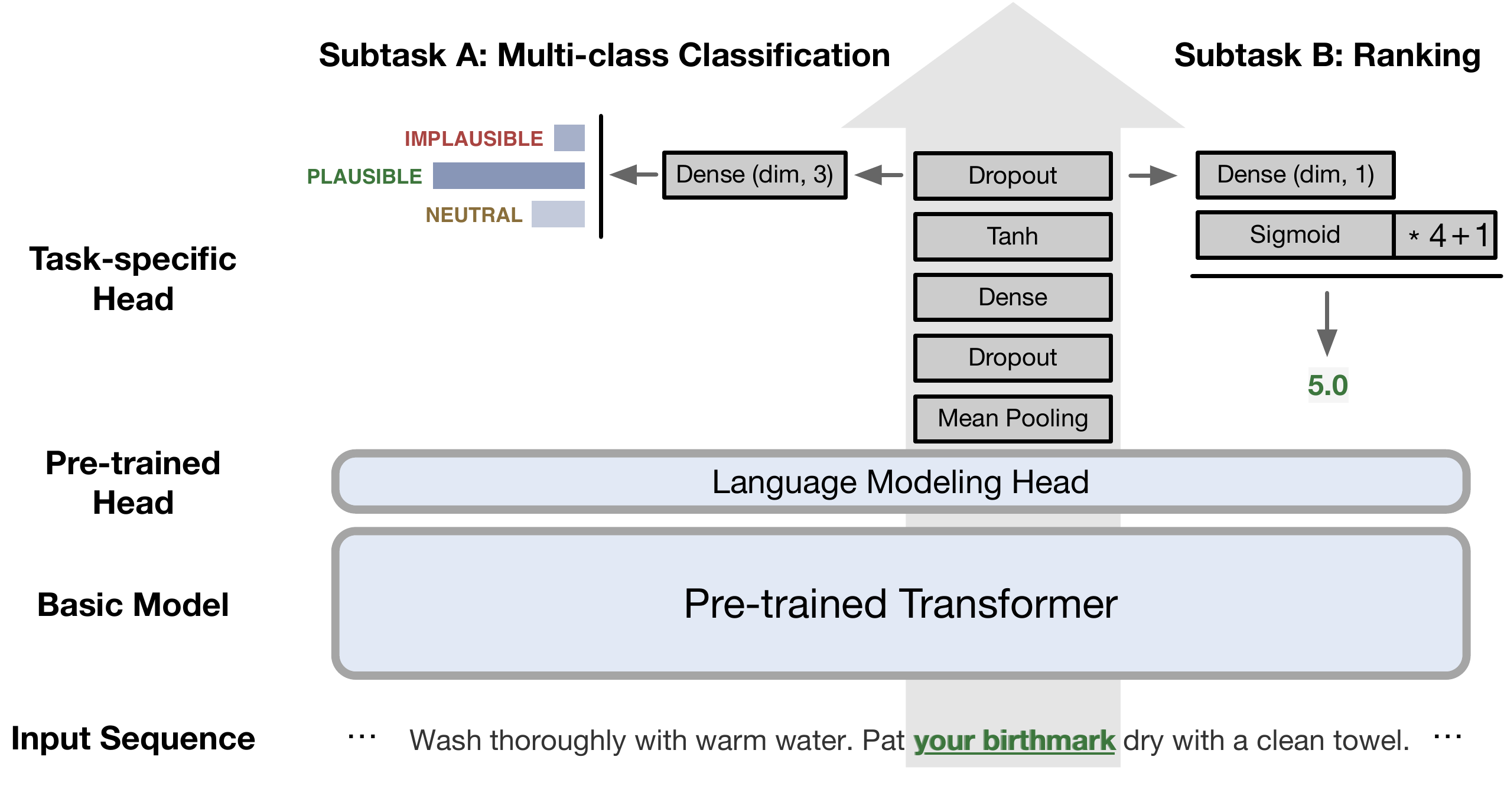}
	\caption{The illustration of our system.}\label{fig:framework}
\end{figure*}

As illustrated in Figure.~\ref{fig:framework}, the framework for both tasks is nearly the same which consists of 4 parts, namely the input, a basic model, a pre-trained head, and a task-specific head.\\

\noindent\textbf{The input sequence}. Each sample is constructed by joining the \textit{Pattern}, \textit{Title}, \textit{Section Header}, \textit{Previous Sentence}, \textit{Target Sentence} and \textit{Follow-up Sentence} in order demonstrated in Figure.~\ref{fig:ex}. Each candidate phase is filled in in its original position in the \textit{Target Sentence}. When modeling the filled target sentence independently, the training set will be 5 times larger than the original since each target phrase has five candidates. \\

\noindent\textbf{Basic Model}. The pre-trained transformer is our starting point. The basic model takes as input the sequence $\tilde{x}$ and outputs the contextual representation of each token as follows:
\begin{equation}
    \mathbf{H}_b=\text{Transformer}(\tilde{x})
\end{equation}
where $\mathbf{H}_b \in \mathbb{R}^{n \times d}$ with $n$ tokens and $d$ dimension.\\

\noindent\textbf{Pre-trained Head}. During the pre-training phase, a language modeling head is appended for a language modeling task. The head is usually discarded in the fine-tuning phase. However, in our experiment, we found better performance can be achieved when reusing the pre-trained head. Formally, the language modeling head takes as the input $\mathbf{H}_b$ and output representations for task-specific head as follows:
\begin{equation}
    \textbf{H}_p = \text{LN}(\text{Act}(\mathbf{H}_b\mathbf{W}_1 +  \mathbf{b}_1))
\end{equation}
where $\mathbf{W}_1 \in \mathbb{R}^{d \times d}, \mathbf{b}_1 \in \mathbb{R}^{d}$ is the weight and bias, $\text{Act}(\cdot)$ and $\text{LN}(\cdot)$ are the activation function and the layernorm layer~\cite{ba2016layer} respectively.\\

\noindent\textbf{Task-specfic Head}. As there are several tokens after tokenizing the target phrase, we apply a mean pooing layer for the hidden states of the target phrase denoted as $\mathbf{H}_{p, i:j}$ as follows:
\begin{equation}
    \mathbf{h}_t = \frac{\sum_i^j \mathbf{H}_{p, i}}{j-i}
\end{equation}
where $\mathbf{h}_t \in \mathbb{R}^{1 \times d}$ is the mean embedding of the target phase, $i, j$ are the start and end token index of the tokenized target phrase respectively.
Then, we sequentially appended a dropout layer, a dense layer with a Tanh activation function and a dropout layer for $\mathbf{h}_t$ as follows:
\begin{equation}
    \mathbf{\tilde{h}}_t = \text{Dropout}(\text{Tanh}(\text{Dropout}(\mathbf{h}_t) \mathbf{W}_2 + \mathbf{b}_2))
\end{equation}
where $\mathbf{\tilde{h}}_t \in \mathbb{R}^{1 \times d}, \mathbf{W}_2 \in \mathbb{R}^{d \times d}, \mathbf{b}_2 \in \mathbb{R}^{d}$ are the enhanced embedding of the target phase, learnable weight and bias respectively.
Finally, the $\mathbf{\tilde{h}}_t$ is transformed to fit the three-class classification task and regression task as follows:
\begin{align}
    \mathbf{\tilde{y}}_c &= \text{Softmax}(\mathbf{\tilde{h}}_t\mathbf{W}_3 + \mathbf{b}_3)\\
    \tilde{y}_r &= \text{Sigmoid}(\mathbf{\tilde{h}}_t\mathbf{W}_4 + \mathbf{b}_4)*4+1
\end{align}
where $\mathbf{\tilde{y}}_c \in \mathbb{R}^{1 \times 3}, \mathbf{W}_3 \in \mathbb{R}^{d \times 3}, \mathbf{b}_3 \in \mathbb{R}^{3}$ are the probabilty distribution, learnable weight and bias for subtask A, and  $\tilde{y}_r \in \mathbb{R}^{1}, \mathbf{W}_3 \in \mathbb{R}^{d \times 1}, \mathbf{b}_3 \in \mathbb{R}^{1}$ are the regression score, learnable weight and bias for subtask B. The Sigmoid function restricts the range of output space between 0 to 1, then we shift the number by multiplying four and adding one. The above method successfully restrict the regression score within the golden score on a scale of 1 to 5.

\subsection{Optimazation Object}\label{sec:opt-obj}
For subtask A, we utilized the cross-entropy loss for multi-class classification as follows:
\begin{equation}
    \mathcal{L}_{ce} = -\sum_i^N \log(\mathbf{\tilde{y}}_c^i[y_c^i])
\end{equation}
where $N$ is the number of training samples, $y_c^i$ is the golden label for $i$-th sample, $\mathbf{\tilde{y}}_c^i[y_c^i]$ means the predicted probability of the golden label.

For subtask B, we used the mean squared error loss for regression as follows:
\begin{equation}
    \mathcal{L}_{reg} = \frac{1}{N} \sum_i^N(\tilde{y}_r^i - y_r^i)^2 
\end{equation}

\subsection{Pattern-aware Ensembling}\label{sec:ensemble}
Ensemble is the commonly used technique where multiple diverse models are trained to predict an outcome, then aggregates the prediction of each model resulting in the final prediction. In our experiment, we observed that the model fine-tuned with different hyper-parameters have different preference on the \textit{Resolved Pattern}\footnote{Descriptions of the resolved pattern can be found in \url{https://competitions.codalab.org/competitions/35210\#participate}}. Thus, we aggregates the prediction of each model seperately based on the performance on a subset split by the given \textit{Resolved Pattern} attribute.

\section{Experiment}
\subsection{Data}
\begin{table}[]
\centering
\resizebox{0.45\textwidth}{!}{%
\begin{tabular}{@{}lccc@{}}
\toprule \toprule
Pattern\textbackslash{}Dataset & Train & Validation & Test \\ \midrule
ADDED COMPOUND                 & 5000  & 625        & 625  \\
FUSED HEAD                     & 4995  & 625        & 625  \\
IMPLICIT REFERENCE             & 4980  & 625        & 625  \\
METONYMIC REFERENCE            & 5000  & 625        & 625  \\ \midrule
Total                          & 19975 & 2500       & 2500 \\ \bottomrule \bottomrule
\end{tabular}%
}
\caption{Data statistics in train, validation, and test set on the different patterns.}
\label{tab:data}
\end{table}
We use the training, validation and test data provided for SemEval 2022 Task 7 without introducing extra data. The data statistic is summarized in Table.~\ref{tab:data} where there is a balanced distribution among different patterns.

\subsection{Experimental Setup}
\begin{table}[]
\centering
\resizebox{0.42\textwidth}{!}{%
\begin{tabular}{@{}lc@{}}
\toprule \toprule
Hyper-parameter     & Model                 \\ \midrule
Dropout             & 0.1                        \\
Warmup Ratio        & 0.1                        \\
Learning Rates      & \{5e-6, 7e-6, 9e-6, 1e-5\} \\
Batch Size          & \{32, 48, 64\}             \\
Weight Decay        & 0.01                       \\
Epoches             & 5                          \\
Learning Rate Decay & Linear                     \\ \bottomrule \bottomrule
\end{tabular}%
}
\caption{Hyper-parameters for fine-tuning on both subtasks.}
\label{tab:hyperparameter}
\end{table}
DeBERTa~\cite{he2020deberta, he2021debertav3}, XLMR~\cite{conneau2019unsupervised} and ERNIE~\cite{sun2021ernie} are used as the pre-trained language model. We fine-tune the models using the AdamW optimizer~\cite{kingma2014adam} with the default hyper-parameter, and additional fine-tuning hyper-parameters are listed in Table.~\ref{tab:hyperparameter}. Experiments are carried out using eight Nvidia A100 GPUs.


\subsection{Evaluation Method}
For subtask A, the evaluation metric is the accuracy score. The model must predict one of the following labels: \{\texttt{IMPLAUSIBLE}, \texttt{NEUTRAL}, \texttt{PLAUSIBLE}\}.

For subtask B, the submission will be scored using Spearman's rank correlation coefficient, which compares the predicted plausibility ranking over all test samples to the gold ranking.

\subsection{Results}

\begin{table}[]
\centering
\resizebox{0.45\textwidth}{!}{%
\begin{tabular}{@{}lllll@{}}
\toprule \toprule
\textbf{Task}   & \multicolumn{2}{l}{\textbf{SubTask-A}} & \multicolumn{2}{l}{\textbf{SubTask-B}} \\ \midrule
  & Dev           & Test                   & Dev           & Test                   \\ \midrule
2nd Place       & -             & 66.10                  & -             & 0.7850                 \\ \midrule
Ensembled Model & 71.08         & 66.50                  & 0.8260        & 0.7950                 \\
+ Pattern-aware & \textbf{75.20}         & \textbf{68.90}         & \textbf{0.8441}        & \textbf{0.8070}        \\ \bottomrule \bottomrule
\end{tabular}%
}
\caption{Performance of models on dev set and official test set.}
\label{tab:official-result}
\end{table}

Our ensembled prediction on test set placed first in the competition, with a 68.9\% accuracy score for subtask A and a 0.8070 Spearman's rank correlation coefficient for subtask B. As shown in Table.~\ref{tab:official-result}. Our system outperforms the second-place system by 2.8 and 2.2 percent points respectively. The organizers predict an upper bound of 77.1\% accuracy score and 0.89 ranking correlation based on the manual annotations. As a result, there's still a lot of room for growth.

\subsection{Ablation Studies}
\begin{table}[]
\centering
\resizebox{0.4\textwidth}{!}{%
\begin{tabular}{@{}lll@{}}
\toprule \toprule
\textbf{\#} & \textbf{Models}   & \textbf{SubTask-A} \\ \midrule
\multicolumn{3}{c}{\textit{MLM-based Models}}        \\ \midrule
1           & XLMR-Large        & 61.14              \\
2           & ERNIE             & 61.73              \\ \midrule
\multicolumn{3}{c}{\textit{RTD-based Models}}        \\ \midrule
3           & DeBERTa-V3-Large  & \textbf{67.25}     \\
4           & \#3 without pre-trained head & 65.96              \\ \bottomrule \bottomrule
\end{tabular}%
}
\caption{Ablation studies on SubTask A with respect to the accuracy score on the dev set. (We reported the mean results with at least three runs.)}
\label{tab:ab-subtaskA}
\end{table}


\begin{figure}[!htb]
	\centering
	\includegraphics[width=0.4\textwidth]{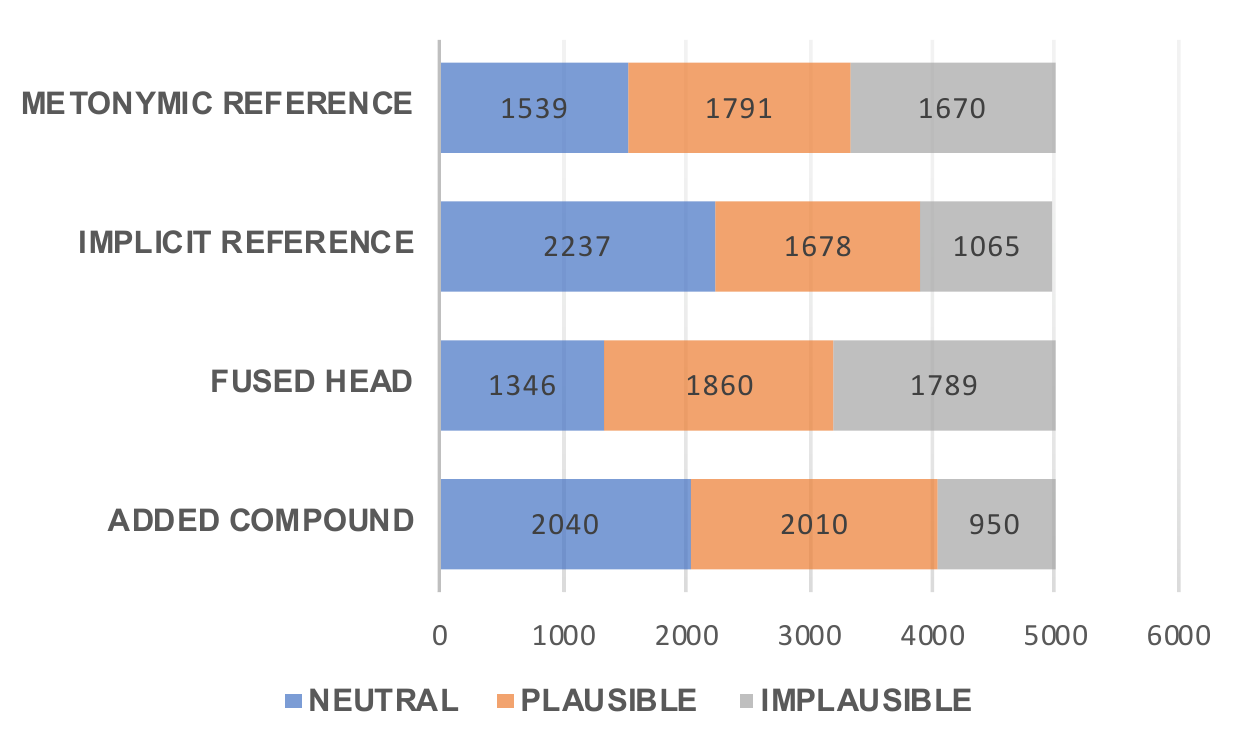}
	\caption{The label distribution of different pattern on training dataset.}\label{fig:dist}
\end{figure}

\begin{table*}[!htb]
\centering
\resizebox{\textwidth}{!}{%
\begin{tabular}{@{}lcccccccc@{}}
\toprule
\textbf{Pattern}                    & \multicolumn{2}{c}{\textbf{IMPLICIT REFERENCE}} & \multicolumn{2}{c}{\textbf{METONYMIC REFERENCE}} & \multicolumn{2}{c}{\textbf{FUSED HEAD}} & \multicolumn{2}{c}{\textbf{ADDED COMPOUND}} \\ \midrule
Hyper-paramters\textbackslash{}Task & SubTask-A              & SubTask-B              & SubTask-A              & SubTask-B               & SubTask-A          & SubTask-B          & SubTask-A            & SubTask-B            \\ \midrule
LR:1e-5, BSZ:32                     & 65.12                  & 0.8321                 & 67.36                  & \textbf{0.8427}         & 67.68              & 0.8400             & 64.64                & \textbf{0.8272}      \\
LR:9e-6, BSZ:32                     & 64.96                  & \textbf{0.8340}        & \textbf{69.60}         & 0.8408                  & \textbf{71.84}     & \textbf{0.8418}    & 63.20                & 0.8251               \\
LR:1e-5, BSZ:64                     & 71.84                  & 0.8286                 & 69.28                  & 0.8382                  & 65.12              & 0.8347             & 68.96                & 0.8134               \\
LR:9e-6, BSZ:64                     & \textbf{72.32}         & 0.8265                 & \textbf{69.60}         & 0.8424                  & 64.96              & 0.8325             & \textbf{69.28}       & 0.8142               \\ \bottomrule
\end{tabular}%
}
\caption{Performance of the DeBERTa-V3-Large with different fine-tuning hyperparameters on the dev set. A model can't win all the subtasks on a subset split by the given pattern attribute. (LR and BSZ are abbreviations for learning rate and batch size.)}
\label{tab:ab-pattern}
\end{table*}
The effectiveness of using a replaced token detection task pre-trained model and recovering the pre-train language modeling head in the task-specific head have been revealed in Table.~\ref{tab:ab-subtaskA}. The hypothesis that utilizing a model pre-trained by a similar task aligned with SemEval-2022 Task 7 contributes a lot is supported by comparing \#1,\#2 and \#3. The performance of the model improved even more after reusing the pre-trained LM head (\#3 and \#4). The assumption is that the hidden states from the pre-trained head contain more information learned during the pre-training phase for distinguishing semantically similar tokens.

The effectiveness of the pattern-aware ensembling has been shown in Table.\ref{tab:official-result}. On subtasks A and B, pattern-aware ensembling outperformed the standard ensemble technique by 2.4 and 1.2 percent points, respectively, compared to the standard ensemble method.

The model trained with different hyper-parameters may perform better on one pattern but not on another, as seen in Table.~\ref{tab:ab-pattern}. For example, on the \texttt{FUSED HEAD} pattern, the model (LR:9e-6, BSZ:32) has the highest accuracy score of 71.84\% but the lowest accuracy scores of 64.96\% and 63.20\% on \texttt{ADDED COMPOUND} and \texttt{IMPLICIT REFERENCE} pattern, respectively. The model (LR:9e-6, BSZ:64), on the other hand, has the lowest score on \texttt{FUSED HEAD} pattern but the best result on \texttt{ADDED COMPOUND} and \texttt{IMPLICIT REFERENCE} pattern. By visualizing the label distribution in Figure.~\ref{fig:dist}, we 
infer that the phenomenon is related to a distribution difference in which \texttt{FUSED HEAD} pattern 
contains the lowest number of the label \textit{NEUTRAL}, and the label \texttt{PLAUSIBLE} dominates the \texttt{ADDED COMPOUND} and \texttt{IMPLICIT REFERENCE} patterns.

\section{Conclusion}
We built a system for identifying plausible clarifications of
implicit and underspecified phrases in instructional texts which is useful for improving the clarification of instructional texts. The system leverages the strength of a replaced token detection pre-trained discriminator and therefore performs extremely well on this shared task with the same goal to distinguish semantically similar tokens. In particular, we proposed a pattern-aware ensembling strategy to aggregate multiple predictions separately based on the pattern when there is a label distribution difference among patterns. On SemEval-2022 Task 7, the system achieved the best performance in both subtasks.

In future work, it's promising to incorporate the replace token detection task in a large-scale pre-trained model with billion, or even trillion parameters. 



\bibliographystyle{acl_natbib}
\bibliography{anthology,acl2021}


\end{document}